\documentclass[conference]{IEEEtran}
\IEEEoverridecommandlockouts
\usepackage{cite}
\usepackage{amsmath,amssymb,amsfonts}
\usepackage{algorithmic}
\usepackage{graphicx}
\usepackage{textcomp}
\usepackage{xcolor}
\usepackage{fontawesome5}
\usepackage{booktabs}
\usepackage{multirow}
\usepackage{pifont}
\usepackage{url}
\usepackage{hyperref}
\usepackage{xurl}
\hypersetup{hidelinks=true}
\usepackage{stackengine}

\usepackage[normalem]{ulem}

\newcommand{\ADDED}[1]{{{#1}}}

\newcommand{\scgp}{\textit{Subject-Conditioned Glucose Prediction}}
\newcommand{\SCGP}{SCGP}

\def\BibTeX{{\rm B\kern-.05em{\sc i\kern-.025em b}\kern-.08em
    T\kern-.1667em\lower.7ex\hbox{E}\kern-.125emX}}
\begin{document}

\title{Subject-Conditioned Glucose Forecasting \\in Type-1 Diabetes}

\author{
\IEEEauthorblockN{
Giorgia Rigamonti,
Mirko Paolo Barbato,
Davide Marelli,
Paolo Napoletano
}
\IEEEauthorblockA{
\textit{Department of Informatics, Systems and Communication} \\
\textit{University of Milano-Bicocca} \\
Milan, Italy \\
\{giorgia.rigamonti, mirko.barbato, davide.marelli, paolo.napoletano\}@unimib.it
}
}

\maketitle

\begin{abstract}

Accurate forecasting of blood glucose concentration is key in the management of Type 1 Diabetes, facilitating early detection of adverse glycemic events and supporting timely therapeutic interventions.
Despite recent advances in glucose prediction, most existing approaches rely on population-level representations or implicit personalization strategies that fail to deliver effective subject-specific forecasts.
In this work, we propose Subject-Conditioned Glucose Prediction (\SCGP), a novel multimodal deep learning architecture conceived for personalized blood glucose prediction. \SCGP{} conditions glucose predictions based on observed glucose data and a compact subject-specific representation learned from contextual information. By explicitly separating subject characterization from glucose dynamics modeling and avoiding early fusion of heterogeneous inputs, the proposed framework effectively captures inter-subject variability while preserving robust and reliable temporal modeling.
Experiments on two state-of-the-art benchmark datasets demonstrate that \SCGP{} consistently improves forecasting performance, enabling reliable detection of adverse glycemic events across multiple prediction horizons, highlighting the benefits of explicit subject conditioning for personalized diabetes management.
\end{abstract}


\section{Introduction}

Type 1 Diabetes (T1D)~\cite{type_1} threatens the lives of an increasing number of patients who need to continuously monitor their blood glucose concentration (BGC) to avoid serious health repercussions. T1D can cause hyper- (above 180 mg/dL) 
and hypo-glycemia (below 70 mg/dL) 
events which respectively cause the abundance and lack of glucose, leading to dangerous situations. To help the management of T1D, the research moved to Continuous Glucose Monitoring (CGM)~\cite{cgm} and the use of an automatic pump to simulate an artificial pancreas~\cite{csii}.

Within this context, substantial effort has focused on forecasting BGC and anticipating adverse glycemic events. The widespread adoption of CGM devices has enabled large-scale longitudinal data collection, fostering the use of machine learning and deep learning for glucose prediction. Despite this progress, accurate forecasting remains challenging due to the complex interaction of subject-specific factors, including physiology, insulin sensitivity, lifestyle habits, diet, physical activity, stress, and sleep~\cite{bent2021engineering}.

Most of the recent studies focus on a general approach that only aims to correctly predict BGCs. These strategies include machine learning techniques such as ARIMA~\cite{arima}, Random Forest, and eXtreme Gradient Boosting (XGBoost)\ADDED{~\cite{alfian2020blood_xgboost}}, and even more complex deep learning techniques\ADDED{~\cite{blood}}. Among these, several models rely exclusively on CGM data, including Informer\ADDED{~\cite{xue2024bgformer}}, GluFormer~\cite{sergazinov2023gluformer}, and TimesNet\ADDED{~\cite{zhang2025blood}}. Other multimodal approaches, such as BG-BERT~\cite{bg-bert}, Bi-GRU~\cite{rigamonti2024improving}, and the Lightweight Sequential Transformer (LST)~\cite{barbatoIJBHI}, incorporate additional physiological signals and lifestyle-related information to enhance predictive performance.

However, many existing approaches do not fully capture the subject-specific characteristics that define individual behavior~\cite{6157604}, limiting the benefits of personalization in improving the accuracy and reliability of future BGC predictions. To address this, recent work has explored personalized modeling strategies that incorporate patient-related information, including static attributes (e.g., age, gender, etc.) and dynamic factors (e.g., sleep, diet, etc.)~\cite{oviedo2017review}. 
%
Most of these efforts focus on Type 2 Diabetes~\cite{deng2021deep}. In contrast, T1D presents distinct challenges, as patients rely entirely on exogenous insulin and require continuous, fine-grained therapy management. This fundamental difference increases the complexity of T1D care and underscores the need for models that explicitly account for its unique treatment dynamics~\cite{podobnik2025metabolic}.

To personalize glucose prediction models, existing studies rely predominantly on strategies such as fine-tuning~\cite{seo2021personalized,deng2021deep} and meta-learning~\cite{langarica2023meta,9813400}, which adapt a population-level model to individual patients. These approaches have proven effective even in the presence of limited subject-specific data, and their benefits have been further confirmed by more recent analyses~\cite{langarica2024deep,lara2025personalized}. However, most of these methods are based primarily on CGM data, and personalization is typically achieved by adjusting model parameters rather than explicitly modeling subject-specific representations.

Although these studies highlight the benefits of personalization, they often \ADDED{underuse} its potential by adapting architectures originally designed for general-purpose prediction to patient-specific modeling, limiting generalizability, particularly in data-scarce settings.
Furthermore, even when multimodal information is considered, additional signals are frequently fused with CGM data at early stages~\cite{rigamonti2026tailoringadverseeventprediction}. Such strategies have been shown to be suboptimal, as they may blur distinct contributions of heterogeneous data sources and hinder the disentanglement of subject-specific traits from glucose dynamics~\cite{stahlschmidt2022multimodal}.

In this paper, we propose \scgp{} (\SCGP{}), a novel multimodal architecture specifically designed for personalized blood glucose forecasting in individuals with T1D. The proposed framework explicitly models subject-specific characteristics and conditions glucose prediction on this information. \SCGP{} is built around two complementary objectives: (i) learning a compact representation of each patient based on personal characteristics and behavioral patterns, and (ii) predicting future blood glucose trajectories. Motivated by the observation that specific contextual modalities—such as basal insulin, bolus insulin, and carbohydrate intake—exhibit strong subject-discriminative power, the model first extracts a subject embedding that captures individual treatment habits and physiological traits. This subject representation is then used to condition the glucose prediction process, coupling subject-specific information with CGM data to improve the accuracy and robustness of future BGC forecasts.

Experiments on two state-of-the-art benchmark datasets demonstrate that \SCGP{} outperforms existing methods in the early detection of hypo- and hyper-glycemic events, while maintaining accurate predictions across different horizons\footnotemark[1].
\footnotetext[1]{\faGithub\ Source code available at
\url{https://github.com/unimib-islab/SCGP}.}

\section{Materials and Methods}

\subsection{Datasets}
In this study, we conducted our experiments on two widely used benchmark datasets for BGC prediction, OhioT1DM~\cite{ohio} and DiaTrend~\cite{diatrend}, which provide multimodal longitudinal data combining CGM with contextual and clinical information.

\ADDED{The OhioT1DM dataset is widely adopted in prior studies on BGC forecasting~\cite{ecai}, and contains data from 12 individuals with T1D collected as part of two challenges (2018 and 2020)}.
For each participant, eight weeks of CGM data are available at five-minute intervals. Beyond glucose measurements, the dataset also includes optional fingerstick glucose readings, detailed records of insulin administration \ADDED{(basal and bolus)}, physiological signals such as heart rate and physical activity, and self-reported information on daily activities and events 
\ADDED{including diet, exercise, sleep, work, stress, and illness}.

The DiaTrend dataset is one of the largest publicly available resources for diabetes research. It includes data from 54 individuals with T1D who participated in the Digital SMD study and the SweetGoals initiative. The dataset contains 27,561 days of CGM data recorded at five-minute intervals and 8,220 days of insulin pump data, including basal insulin for 17 participants. It also provides detailed records of bolus insulin, carbohydrate intake, pump settings, and comprehensive demographic and clinical information such as age, sex, race, and hemoglobin A1C (HbA1c) levels. 

\subsection{Data Pre-processing}

To ensure consistency, we applied a standard pre-processing pipeline to both datasets. First, all input features were synchronized with BGC readings following~\cite{bg-bert, rigamonti2024improving}, aligning each timestamp with subsequent observations within a maximum delay of four minutes to prevent information leakage. Next, we constructed time series from continuous glucose monitoring intervals. Linear interpolation enforced a uniform five-minute sampling interval and filled occasional CGM gaps. When gaps exceeded 20 minutes (more than five consecutive samples), we split the sequence rather than imputing values, creating a new time series.
Following~\cite{butt2023feature}, bolus insulin was transformed via an exponential decay function to estimate insulin on board (IOB), while carbohydrate intake was modeled with a custom absorption function to capture time-varying post-ingestion dynamics. \ADDED{Basal insulin, being a quasi-continuous signal, was represented as a step-wise constant time series, with values propagated forward until updated. For the OhioT1DM dataset, which also includes temporary basal rates, these were used when present in place of the standard basal rate.} Finally, all features were scaled using min–max normalization.

\subsection{Data Split}
We adopted a temporal data split, training on past data, validating on intermediate data, and testing on future data, which is standard practice in time-series forecasting~\cite{barbatoIJBHI}.
For OhioT1DM, we used the original train-test partition and reserved the last 20\% of the training set for validation~\cite{ohio}. For DiaTrend, we used a 64/16/20 split for training, validation, and testing, in line with prior works~\cite{bg-bert, barbatoIJBHI}.

\subsection{Handling Class Imbalance}
Both datasets exhibit a strong imbalance between normal and adverse BGC. In OhioT1DM, hypoglycemic and hyperglycemic readings account for 3.28\% and 8.29\% of samples, respectively. Similarly, in DiaTrend, hypoglycemia represents 1.45\% of readings, while hyperglycemia accounts for 10.42\%.

To mitigate this imbalance, we applied the Synthetic Minority Over-sampling Technique (SMOTE)~\cite{chawla2002smote} exclusively to the training sets, following established practices~\cite{bg-bert, rigamonti2024improving}. 

\subsection{\SCGP{} Architecture}

\begin{figure}[tb]
    \centering
    \includegraphics[width=\linewidth]{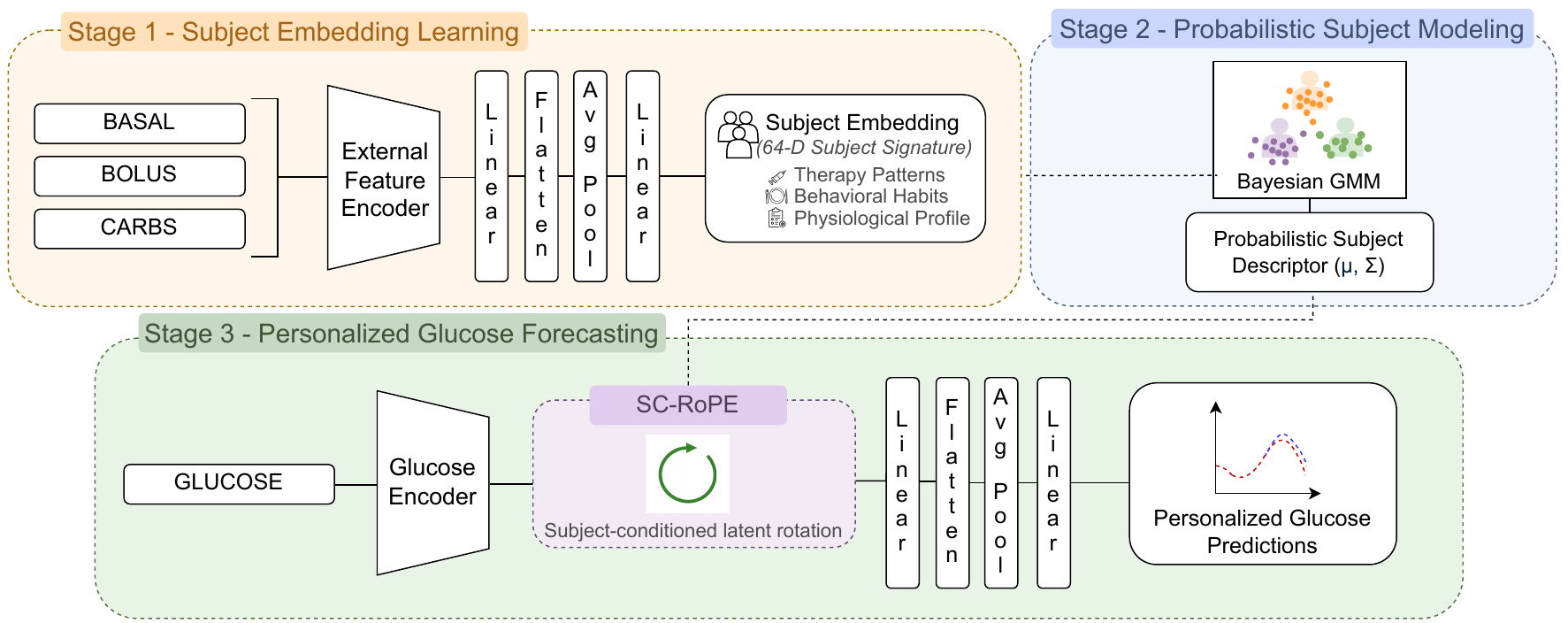}
    \caption{Architecture of the proposed \scgp{} framework. Subject embeddings learned from contextual signals are probabilistically modeled via a Bayesian GMM and injected into the glucose forecasting network through SC-RoPE.}
    \label{fig:model_architecture}
\end{figure}

The proposed architecture is composed of distinct but tightly integrated components, each designed to address a specific aspect of personalized BGC forecasting: (1) subject characterization from contextual signals, (2) probabilistic modeling of subject variability, (3) subject-conditioned modulation of glucose dynamics, and personalized regression of future BGC trajectories (Figure~\ref{fig:model_architecture}).

\subsubsection{Subject Embedding Learning}

In the first stage of the architecture, we aim to learn a compact and informative representation of each user’s typical behavior and treatment management. Rather than operating on raw signals, we summarize this information into a low-dimensional, static embedding that serves as a personalized behavioral signature. This embedding captures individual differences in eating habits and insulin administration, enabling later stages to make user-specific predictions instead of relying on a one-size-fits-all approach.

To construct this representation, we focus exclusively on carbohydrate intake, bolus insulin, and basal insulin. These are processed by a dedicated recurrent encoder (the External Feature Encoder in Figure~\ref{fig:model_architecture}) composed of two gated recurrent unit (GRU) layers. A bidirectional GRU with hidden size 128 first captures past and future temporal context, producing a 256-dimensional output at each time step. This is followed by a unidirectional GRU with hidden size 256 to further refine temporal features, after which dropout with rate 0.4 is applied.

The resulting sequence is projected into a 512-dimensional space through a linear layer, followed by a LeakyReLU activation ($\alpha = 0.01$) and an additional dropout layer. A subsequent fully connected layer reduces the features to 128 dimensions per time step. The temporal dimension is then compressed via one-dimensional average pooling with kernel and stride of 2. The pooled features are flattened and passed through a final linear layer to produce a 64-dimensional embedding.

The embedding is trained using an ArcFace loss~\cite{deng2019arcface}, which encourages compact intra-subject clusters and greater inter-subject separation. This results in stable subject representations that reduce ambiguity between dissimilar users while preserving meaningful similarities among related behavioral patterns.

\subsubsection{Probabilistic Subject Modeling with BGMM}

Once the subject-specific embeddings are obtained from the classifier, we model their distribution in the latent space using a Bayesian Gaussian Mixture Model (BGMM) to derive a compact, probabilistic subject descriptor. Let $\mathbf{z}_{ij} \in \mathbb{R}^{d}$ denote the $j$-th embedding of subject $i$ (with $d=64$ in our implementation). The BGMM is used to capture the underlying structure of the population by identifying groups of subjects with similar behavioral and treatment patterns in a soft, probabilistic manner, rather than assigning them to rigid, deterministic clusters.

Specifically, the BGMM models the distribution of the embeddings as a mixture of $K$ Gaussian components:
\begin{equation}
p(\mathbf{z}_{ij}) = \sum\nolimits^{K}_{k=1} \pi_k \, \mathcal{N}(\mathbf{z}_{ij} \mid \boldsymbol{\mu}_k, \boldsymbol{\Sigma}_k),
\end{equation}
where the mixture weights $\boldsymbol{\pi} = (\pi_1,\dots,\pi_K)$ 
are assigned a $\mathrm{Dirichlet}(\boldsymbol{\alpha})$ prior. For each component $k$, the mean and covariance are assigned a conjugate
Normal--Inverse--Wishart prior:
\begin{equation}
\boldsymbol{\Sigma}_k \sim \mathcal{IW}(\mathbf{W}_0, \nu_0), \qquad
\boldsymbol{\mu}_k \mid \boldsymbol{\Sigma}_k \sim
\mathcal{N}(\mathbf{m}_0, \beta_0^{-1}\boldsymbol{\Sigma}_k).
\end{equation}
%

Each component represents a latent subgroup of subjects with similar characteristics in the embedding space.

For each embedding $\mathbf{z}_{i,j}$, the BGMM provides a responsibility vector which represents how much each mixture component explains each embedding:
\begin{equation}
\boldsymbol{\gamma}_{i,j} = [\gamma_{i,j}^{(1)}, \dots, \gamma_{i,j}^{(K)}],
\end{equation}
where $\gamma_{i,j}^{(k)} = q(k \mid \mathbf{z}_{i,j})$, obtained via variational Bayesian inference, denotes the approximate posterior probability that the embedding belongs to component $k$.

To obtain a subject-level representation, we aggregate these responsibilities across all embeddings of subject $i$, $\bar{\boldsymbol{\gamma}}_i = \frac{1}{N_i} \sum_{j=1}^{N_i} \boldsymbol{\gamma}_{i,j}$, and we compute a composite Gaussian distribution with mean $\boldsymbol{\mu}_i$ 
and covariance $\boldsymbol{\Sigma}_i$:
\begin{equation}
\boldsymbol{\mu}_i = \sum\nolimits_{k=1}^{K} \bar{\gamma}_i^{(k)} \, \boldsymbol{\mu}_k,
\end{equation}
\begin{equation}
\boldsymbol{\Sigma}_i = \sum\nolimits_{k=1}^{K} \bar{\gamma}_i^{(k)} \left( 
\boldsymbol{\Sigma}_k + (\boldsymbol{\mu}_k - \boldsymbol{\mu}_i)(\boldsymbol{\mu}_k - \boldsymbol{\mu}_i)^\top 
\right).
\end{equation}

\noindent The final Gaussian-based subject embedding is obtained by concatenating the mean vector and the diagonal of the covariance matrix:
\begin{equation}
\mathbf{g}_i = \left[ \boldsymbol{\mu}_i \, \| \, \mathrm{diag}(\boldsymbol{\Sigma}_i) \right] \in \mathbb{R}^{2d},
\end{equation}
which provides a compact statistical summary that captures both the central tendency and the variability of subject $i$’s contextual patterns in the latent space.

\subsubsection{Personalized Glucose Regression Module}
Building upon the subject representations learned in the previous stages, the personalized glucose regression module translates subject-specific behavioral information into individualized BGC trajectory predictions by jointly modeling temporal glucose dynamics and subject characteristics. Rather than treating personalization as an external conditioning signal, subject information is directly embedded into the temporal glucose representation through a subject-conditioned rotary feature modulation mechanism.

The input to the regression module is a univariate BGC sequence, which is encoded by a recurrent glucose encoder (same as the one used for Subject Embedding but without the last pooling and projection layers) to extract 128-dimensional temporal features.

Personalization is achieved by conditioning these temporal representations on the subject-specific behavioral embedding learned by the subject embedding learning module. Specifically, we use the 64-dimensional embedding obtained from the subject embedding learning stage, which summarizes the individual’s typical eating habits and insulin management strategies. This embedding is first $\ell_2$-normalized and then linearly projected to a $D/2$-dimensional vector to match half of the glucose feature dimension. The projected vector implicitly parameterizes a set of rotation frequencies that depend on both subject identity and temporal position.

Subject conditioning is implemented through a Subject-Conditioned Rotary Positional Embedding (SC-RoPE) mechanism. Formally, let $\mathbf{f}_t \in \mathbb{R}^{D}$ denote the glucose representation at time step $t$ for a given subject $i$, and let $\mathbf{e}_i \in \mathbb{R}^{D/2}$ be the projected and normalized subject embedding. Given the temporal position $m_t \in \{0, \dots, T-1\}$, SC-RoPE computes a rotation angle
\begin{equation}
\boldsymbol{\theta}_{i,t} = m_t \cdot \mathbf{e}_i .
\end{equation}

The glucose representation is then decomposed into even and odd components,
$\mathbf{f}_t^{\text{even}}$ and $\mathbf{f}_t^{\text{odd}}$, and rotated as:
\begin{align}
\tilde{\mathbf{f}}_t^{\text{even}} &= 
\mathbf{f}_t^{\text{even}} \odot \cos(\boldsymbol{\theta}_{i,t})
- \mathbf{f}_t^{\text{odd}} \odot \sin(\boldsymbol{\theta}_{i,t}), \\
\tilde{\mathbf{f}}_t^{\text{odd}} &=
\mathbf{f}_t^{\text{even}} \odot \sin(\boldsymbol{\theta}_{i,t})
+ \mathbf{f}_t^{\text{odd}} \odot \cos(\boldsymbol{\theta}_{i,t}),
\end{align}
where $\odot$ denotes element-wise multiplication. The final rotated representation $\tilde{\mathbf{f}}_t$ is obtained by interleaving $\tilde{\mathbf{f}}_t^{\text{even}}$ and $\tilde{\mathbf{f}}_t^{\text{odd}}$ along the feature dimension.

This subject-conditioned rotation aligns the temporal glucose encoding with the individual behavioral signature, allowing the regression model to adapt its internal representation based on subject-specific characteristics. In practice, SC-RoPE injects subject information directly into the temporal glucose dynamics while preserving their sequential structure, thereby enabling subject-aware forecasting.

Together with the probabilistic subject descriptors derived from the BGMM, SC-RoPE enables personalization at complementary levels: the BGMM structures the population-level latent space and captures uncertainty in subject grouping, while SC-RoPE directly modulates the temporal glucose representation at the individual level during inference.

After subject-specific conditioning, the glucose representation is passed to the regression head, which applies a linear layer to reduce the channel dimension to 64, followed by one-dimensional average pooling (kernel and stride of 2). The pooled features are flattened and processed by a final linear layer to produce the predicted glucose trajectory.

\subsection{Training Procedure}
Training consists of three sequential stages. We first learn subject-specific behavioral embeddings through the classification network based on non-CGM contextual features, which are used as fixed subject descriptors in the subsequent regression stages.
We then perform subject-agnostic pre-training of the personalized glucose forecasting model 
to learn general, subject-independent glucose dynamics.
During this phase, the glucose forecasting architecture and the SC-RoPE mechanism remain unchanged; however, subject conditioning is explicitly disabled by replacing the subject embedding input with a small-magnitude Gaussian noise vector.
The pretrained model obtained in this stage is used to initialize the subsequent \ADDED{Leave-One-Subject-Out Cross-Validation (LOSOCV)} experiments on both OhioT1DM and DiaTrend.

In the second stage, we adopt a LOSOCV protocol, where for each fold, one subject is held out for testing while the remaining subjects are used for training and validation. In this stage, personalization is achieved exclusively through subject-aware conditioning: subject-level embeddings obtained from the external-feature encoder are aggregated via a BGMM and injected into the regression model through SC-RoPE, while the glucose encoder is kept frozen to preserve subject-independent glucose temporal dynamics.

Finally, in the fine-tuning stage, the model obtained from LOSOCV is further adapted to each subject using their own data, allowing subject-specific parameter adaptation.

Across all experiments, the subject-agnostic pretraining stage always uses the same 37 DiaTrend subjects from the SweetGoals study, while the subsequent LOSOCV and fine-tuning stages are performed using either the OhioT1DM subjects or the 17 DiaTrend subjects from the Digital SMD study, depending on the experimental setup.

\subsection{Composite Loss for Personalized Glucose Forecasting}

During LOSOCV and subject-specific fine-tuning, the regression model is optimized using a composite loss designed to jointly address predictive accuracy, temporal alignment, robustness, and clinical relevance. The overall objective combines four complementary components:
\begin{equation}
\mathcal{L} = \mathcal{L}_{q}
+ \lambda_s \mathcal{L}_{s}
+ \lambda_e \mathcal{L}_{e}
+ \lambda_{\ell} \mathcal{L}_{\ell},
\end{equation}
where $\lambda_s$, $\lambda_e$, and $\lambda_{\ell}$ control the relative contribution of each term 
($\lambda_s=0.4$, $\lambda_e=0.6$, and $\lambda_{\ell}=0.07$ during LOSOCV and fine-tuning).

\paragraph{Multi-quantile loss $\mathcal{L}_{q}$}
To model predictive uncertainty, the model outputs multiple quantiles of the future glucose trajectory. A standard quantile loss is applied to the predicted quantiles $\mathcal{Q}=\{0.1,0.5,0.9\}$:
\begin{equation}
\mathcal{L}_{q} =
\frac{1}{|\mathcal{Q}|}\sum\nolimits_{q \in \mathcal{Q}}
\mathbb{E}\!\left[\max\!\big(q e,\,(q-1)e\big)\right],
\qquad
e = y - \hat{y}^{(q)} .
\end{equation}
where $y$ is the ground-truth glucose value and $\hat{y}^{(q)}$ indicates the corresponding predicted glucose value at quantile $q$.

\paragraph{Robustness and smoothness regularization $\mathcal{L}_{s}$}
To improve stability, additional regularization terms are applied to the median quantile prediction. 
Following the formulation adopted in~\cite{bg-bert}, this regularization consists of two complementary components: a robustness term that mitigates the influence of outliers through a shrinkage penalty on the prediction error, and a smoothness term that enforces temporal consistency by penalizing discrepancies between consecutive prediction differences using a mean squared error (MSE) on first-order temporal differences.
The relative contribution of these terms is automatically balanced during training using CoV-weighting.

\paragraph{Event-aware penalty $\mathcal{L}_{e}$}
Clinical relevance is incorporated through an event-aware penalty applied to the median prediction. Target values are categorized into hypoglycemia ($y<70$ mg/dL), normoglycemia ($70\leq y \leq 250$ mg/dL), and severe hyperglycemia ($y>250$ mg/dL). The loss penalizes clinically undesirable behaviors using a fixed safety margin $\delta$:
\begin{equation}
\mathcal{L}_{e} =
\begin{cases}
\max(0, \hat{y}-y-\delta), & y<70 \\
\max(0, | \hat{y}-y |-\delta), & 70 \le y \le 250 \\
\max(0, y-\hat{y}-\delta), & y>250 .
\end{cases}
\end{equation}

\paragraph{Lag-aware alignment loss $\mathcal{L}_{\ell}$}
To reduce systematic temporal delays between predictions and reference glucose trajectories, we introduce a lag-aware alignment loss applied to the median prediction. For each prediction window, the MSE is evaluated over a set of non-negative temporal shifts $k \in [0, K]$, where $K$ denotes the maximum allowed temporal misalignment. In our experiments, we set $K=3$ time steps, defining the temporal tolerance. These errors are then used to compute a softmin distribution over the possible shifts, assigning higher weight to shifts associated with lower reconstruction error.

Rather than penalizing the reconstruction error directly, the loss minimizes the expected temporal lag under this softmin distribution, encouraging predictions that are well aligned and anticipate future glucose changes:
\begin{equation}
\mathcal{L}_{\ell} = \mathbb{E}_{k \sim p(k)}[k],
\end{equation}
where $p(k)$ denotes the softmin-normalized distribution derived from the shift-dependent mean squared errors. This formulation promotes earlier and better-aligned forecasts without enforcing a fixed temporal offset.

\section{Results and Discussion}
All experiments were conducted on both datasets under two standard prediction horizon settings. 
A 30-minute horizon (PH = 30 min) was modeled using a 150-minute temporal window, comprising 120 minutes (24 timestamps) for observation and 30 minutes (6 timestamps) for prediction. Similarly, a 60-minute horizon (PH = 60 min) was implemented using a 300-minute window, with 240 minutes (48 timestamps) allocated to observation and 60 minutes (12 timestamps) to prediction.

\subsection{Evaluation Metrics} 
For a comprehensive assessment of the proposed model, we consider two complementary perspectives: an analytical evaluation, aimed at quantifying prediction accuracy and event detection capability, and a clinical evaluation, focused on the safety and reliability of the forecasts in real-world scenarios.

\subsubsection{Analytical Evaluation Metrics} 
We evaluate all experiments using standard BGC prediction metrics~\cite{bg-bert,rigamonti2024improving}: Root Mean Square Error (RMSE), Time Gain (TG), and sensitivity/specificity for hyper- and hypoglycemia.



The Time Gain (TG) measures prediction anticipation as $\operatorname{TG}(\mathbf{y}, \mathbf{\hat{y}}) = PH - \operatorname{delay}(\mathbf{y}, \mathbf{\hat{y}})$, where $PH$ is the prediction horizon and $\operatorname{delay}(\mathbf{y}, \mathbf{\hat{y}}) = \mathrm{argmin}k \sum{i=1}^L (y_i - \hat{y}_{i-k})^2$ is the temporal shift minimizing the squared error between predicted and reference trajectories.


Event detection is assessed via sensitivity and specificity: sensitivity (Hypo/Hyper Sen) measures correct detection of adverse events, while specificity (Hypo/Hyper Spec) reflects correct identification of non-adverse conditions, limiting false alarms.
Finally, the False Alarm Rates (FARs) are computed as $1 - \text{Specificity}$.

\subsubsection{Clinical Evaluation Metrics}


To evaluate the clinical reliability of BGC forecasts, we use Clarke’s Error Grid Analysis (EGA)~\cite{ega}. This widely adopted, clinically grounded method assesses the potential risk of prediction errors by comparing predicted and reference BGC values on a grid divided into five zones (A–E).

Zones A and B denote clinically acceptable predictions that would lead to correct or benign treatment, whereas zones C and D reflect increasingly unsafe errors that may prompt inappropriate actions. Zone E corresponds to the most dangerous errors, with the potential for severe clinical consequences.

\subsection{Effect of Personalization via Subject Conditioning}

\begin{table*}[tb]
\centering
\caption{Patient-Independent vs. Patient-Specific models Results\textsuperscript{*}}
\label{tab:results}
\resizebox{\textwidth}{!}{%
\begin{tabular}{llccccccccccccccc}
\toprule
 & & & \multicolumn{6}{c}{PH = 30 mins} & \multicolumn{6}{c}{PH = 60 mins} \\  
\cmidrule(lr){4-9} \cmidrule(lr){10-15}
Dataset & Model & Patient- 
& RMSE & TG & Hyper Sen & Hypo Sen & Hyper Spec & Hypo Spec
& RMSE & TG & Hyper Sen & Hypo Sen & Hyper Spec & Hypo Spec \\ 
 & & Specific
& (mg/dL) ↓ & (mins) ↑ & (\%) ↑ & (\%) ↑ & (\%) ↑ & (\%) ↑
& (mg/dL) ↓ & (mins) ↑ & (\%) ↑ & (\%) ↑ & (\%) ↑ & (\%) ↑\\
\midrule
\hfil\multirow{6}{*}{\rotatebox[origin=c]{90}{OhioT1DM}}\hfill
& CNN       & --  
                & 12.97 $\pm$ 1.92 & 15.70 $\pm$ 0.94 & 83.03 $\pm$ 10.80 & 70.26 $\pm$ 15.34 & 99.17 $\pm$ 0.44 & 99.14 $\pm$ 0.46
                & 22.01 $\pm$ 2.74 & 28.15 $\pm$ 3.54 & 65.92 $\pm$ 13.85 & 45.01 $\pm$ 17.40 & 98.99 $\pm$ 0.48 & 98.89 $\pm$ 0.64\\
& CNN       & \checkmark  
                & 12.85 $\pm$ 1.89 & 15.99 $\pm$ 1.09 & 83.31 $\pm$ 11.17 & 66.45 $\pm$ 19.16 & 99.12 $\pm$ 0.63 & \textbf{99.16} $\pm$ 0.43
                & \textbf{21.35} $\pm$ 2.68 & 28.76 $\pm$ 2.66 & 65.64 $\pm$ 14.97 & 43.47 $\pm$ 21.86 & 98.83 $\pm$ 0.88 & \textbf{98.98} $\pm$ 0.75\\
& Bi-GRU    & --  
                & 12.85 $\pm$ 1.94 & 16.91 $\pm$ 0.95 & 81.63 $\pm$ 10.66 & 73.93 $\pm$ 16.19 & \textbf{99.34} $\pm$ 0.45 & 99.08 $\pm$ 0.41
                & 22.89 $\pm$ 2.75 & 32.07 $\pm$ 3.18 & 59.58 $\pm$ 16.77 & 48.13 $\pm$ 16.96 & \textbf{99.39}	$\pm$ 0.34 & 98.82 $\pm$ 0.61\\ 
& Bi-GRU    & \checkmark
                & 13.66 $\pm$ 3.16 & 18.41 $\pm$ 1.64 & 83.24 $\pm$ 9.86 & 80.37 $\pm$ 14.56 & 99.11 $\pm$ 0.98 & 98.59 $\pm$ 0.77
                & 23.09 $\pm$ 3.91 & \textbf{33.63} $\pm$ 2.87 & 66.29 $\pm$ 13.62 & 53.78 $\pm$ 21.14 & 98.93 $\pm$ 0.98 & 98.26 $\pm$ 1.07\\ 	
& \SCGP{} & -- 
                & \textbf{12.46} $\pm$ 1.81 & 17.91 $\pm$ 1.41 & 84.21 $\pm$ 10.67 & 79.45 $\pm$ 20.58 & 99.17 $\pm$ 0.49 & 98.75 $\pm$ 0.67
                & 21.51 $\pm$ 2.49 & 32.09 $\pm$ 4.19 & 66.97 $\pm$ 14.94 & 57.97 $\pm$ 20.79 & 99.08 $\pm$ 0.53 & 98.53 $\pm$ 0.95\\
& \SCGP{} & \checkmark
                & 12.86 $\pm$ 2.15 & \textbf{18.86} $\pm$ 1.05 & \textbf{85.46} $\pm$ 11.17 & \textbf{81.97} $\pm$ 13.58 & 98.93 $\pm$ 0.94 & 98.47 $\pm$ 0.73 
                & 21.73 $\pm$ 3.22 & 32.85 $\pm$ 3.00 & \textbf{70.15} $\pm$ 16.17 & \textbf{63.98} $\pm$ 22.20 & 98.57	$\pm$ 1.17 & 97.92 $\pm$ 1.12\\
\midrule
\hfil\multirow{6}{*}{\rotatebox[origin=c]{90}{DiaTrend}}\hfill
& CNN       & --    
                & 14.97 $\pm$ 3.29 & 15.18 $\pm$ 1.63 & 76.95 $\pm$ 20.21 & 48.33 $\pm$ 9.43 & 98.78 $\pm$ 1.11 & 99.34 $\pm$ 1.15
                & 24.30 $\pm$ 4.71 & 28.09 $\pm$ 3.43 & \textbf{63.37} $\pm$ 19.62 & 24.95 $\pm$ 5.76 & 98.38 $\pm$ 1.65 & \textbf{99.51} $\pm$ 0.97\\
& CNN       & \checkmark  
                & 14.70 $\pm$ 3.21 & 15.46 $\pm$ 1.73 & 75.90 $\pm$ 22.01 & 47.39 $\pm$ 15.93 & 98.79 $\pm$ 1.31 & \textbf{99.35} $\pm$ 1.29
                & \textbf{23.74} $\pm$ 4.66 & 28.61 $\pm$ 3.26 & 61.27 $\pm$ 22.52 & 26.84 $\pm$ 15.40 & 98.28 $\pm$ 2.24 & 99.44 $\pm$ 1.07\\
& Bi-GRU    & --
                & 14.80 $\pm$ 3.13 & 16.31 $\pm$ 1.67 & 74.51 $\pm$ 20.04 & 51.93 $\pm$ 12.32 & \textbf{99.11} $\pm$ 0.80 & 99.20 $\pm$ 1.56
                & 24.88 $\pm$ 5.11 & 31.01 $\pm$ 3.85 & 58.56 $\pm$ 19.73 & 24.20 $\pm$ 4.79 & \textbf{98.84} $\pm$ 1.20 & 99.24 $\pm$ 2.22\\
& Bi-GRU    & \checkmark 
                & 15.04 $\pm$ 3.30 & 17.33 $\pm$ 1.74 & 73.20 $\pm$ 22.90 & 60.10 $\pm$ 17.22 & 98.94 $\pm$ 1.20 & 98.32 $\pm$ 4.41
                & 25.14 $\pm$ 5.42 & 32.48 $\pm$ 3.94 & 59.98 $\pm$ 22.65 & 36.39 $\pm$ 21.71 & 98.37 $\pm$ 2.23 & 98.88 $\pm$ 1.67\\	
& \SCGP{} & -- 
                & \textbf{14.53} $\pm$ 2.84 & 16.34 $\pm$ 2.05 & 76.23 $\pm$ 20.29 & 53.66 $\pm$ 18.76 & 98.91 $\pm$ 1.10 & 99.19 $\pm$ 1.35
                & 25.16 $\pm$ 4.16 & \textbf{34.13} $\pm$ 4.61 & 57.29 $\pm$ 24.46 & 25.60 $\pm$ 14.08 & 98.75	$\pm$ 1.19 & 99.46 $\pm$ 0.78\\
& \SCGP{} & \checkmark
                 & 14.76 $\pm$ 3.02 & \textbf{17.48} $\pm$ 1.70 & \textbf{78.18} $\pm$ 20.02 & \textbf{70.43} $\pm$ 12.69 & 98.78 $\pm$ 1.14 & 97.74 $\pm$ 5.77
                 & 24.10 $\pm$ 4.42 & 31.88 $\pm$ 4.50 & 62.69 $\pm$ 23.73 & \textbf{43.77} $\pm$ 20.65 & 98.41	$\pm$ 1.97 & 98.36 $\pm$ 2.65\\

\bottomrule
\end{tabular}}
\\
 \vspace{0.3em}
 \raggedright{
 \centering
 \scriptsize{\textsuperscript{*} In \textbf{bold} the best score; Values are reported as mean  $\pm$ standard deviation across patients.}\\
 }
\end{table*}

Based on the evaluation metrics described above, Table~\ref{tab:results} reports the performance of patient-independent and patient-specific models on the OhioT1DM and DiaTrend datasets for prediction horizons of 30 and 60 minutes. 

We compare the proposed approach against two strong personalized baselines representing complementary modeling paradigms. The first is a CNN-based model derived from Deng et al.~\cite{deng2021deep}, which relies solely on CGM data and combines TimeGAN-based augmentation with a convolutional architecture. Owing to its strong performance and widespread adoption, it serves as a reference for CGM-only personalized glucose prediction. For fairness, we adapt it to predict glucose changes between time steps $t$ and $t+PH$, consistent with our formulation.
The second baseline is a Bi-GRU-based personalized model derived from~\cite{rigamonti2026tailoringadverseeventprediction}, which employs a recurrent architecture and a two-stage training strategy consisting of patient-independent training followed by subject-specific fine-tuning. Although originally designed for multimodal inputs, we evaluate this model using CGM data only to isolate the effects of temporal modeling and personalization.

This setup reflects a realistic deployment scenario. While the proposed \SCGP{} model leverages multimodal data during an initial observation phase to learn a subject-specific embedding, glucose forecasting at inference time relies exclusively on CGM measurements. Restricting all methods to CGM input only during prediction ensures a fair comparison and closely matches real-world conditions.

Results in Table~\ref{tab:results} show a consistent advantage of patient-specific modeling across both datasets and prediction horizons. Personalized models achieve lower RMSE and higher Time Gain than patient-independent counterparts, indicating improved accuracy and earlier anticipation of adverse events.

\SCGP{} demonstrates the most stable and balanced performance across metrics. In both datasets, the patient-specific configuration achieves competitive RMSE while consistently improving hypoglycemic sensitivity without degrading specificity. Unlike standard CNN- or RNN-based personalization approaches that implicitly entangle subject information with glucose dynamics through fine-tuning, the proposed framework explicitly decouples subject representation learning from temporal glucose modeling. This is particularly relevant clinically, as hypoglycemic events are rare but critical.

These benefits are more pronounced at the longer prediction horizon (PH = 60 min), where forecasting is more challenging. In this setting, \SCGP{} maintains stable accuracy and Time Gain while providing a marked improvement in hypoglycemia sensitivity, suggesting that subject-aware learning helps preserve clinically relevant information over longer horizons.

Finally, we analyze the FAR to assess practical implications in terms of alarm fatigue. \SCGP{} models exhibit low FAR, with patient-specific configurations showing only marginal increases (+0.1\% at PH = 30, +0.4\% at PH = 60), reflecting the expected trade-off between sensitivity and false alarms. Compared to the baselines, \SCGP{} also shows more stable FAR across settings, indicating that improved detection does not substantially increase alarm burden.

\subsection{Clinical Evaluation via Clarke's Error Grid Analysis}
\ADDED{Across datasets and PHs, the vast majority of predictions fall within zones A and B of the Clarke Error Grid, indicating clinically acceptable performance.

At PH = 30 min, zone A accounts for approximately 93--96\% of predictions, with most samples in zone B. Clinically unsafe zones (C--E) are extremely rare: zone D is below 0.5\%, while zones C and E are near zero across configurations.

As the PH increases to 60 min, the proportion of samples in zone A decreases to approximately 83--87\%, with a corresponding shift toward zone B ($\approx$12--16\%), as expected. Nevertheless, the combined percentage of zones A and B consistently exceeds 98\%, indicating that most prediction errors would still lead to correct or benign treatment decisions.

Patient-specific models show EGA distributions comparable to patient-independent ones, with a slight reduction in critical zones. Specifically, zone D decreases from 0.8\% to 0.7\% on OhioT1DM and from 1.6\% to 1.0\% on DiaTrend.
Predictions falling into zone E are virtually absent in all experiments ($\leq$0.02\%), while zone C remains below 0.1\%, confirming that personalization preserves clinical reliability without introducing potentially dangerous errors.}

\subsection{Impact of Multimodality on Subject-Specific Modeling}
To further motivate our design, we perform a feature ablation study on both datasets (Table~\ref{tab:abl_ext_features}) to assess how combinations of external physiological inputs (i.e., basal insulin, bolus insulin, and carbohydrate intake) contribute to subject discrimination. 
Results clearly indicate that these features provide information that supports subject discriminability. Among them, basal insulin shows strong discerning power. The use of all three features achieves the best results across both datasets.


These findings motivate the use of a dedicated encoder for contextual features, separate from the CGM encoder that models glucose dynamics as proposed in our multitask architecture. 
This design reflects the distinct roles of the two modalities: external signals primarily capture subject-specific physiological and behavioral traits, whereas CGM encodes the temporal evolution of glucose levels.

The proposed architecture leverages all three features during an initial observation phase to learn a compact subject-specific embedding. 
This embedding captures individual characteristics that are not fully observable from CGM alone, as evidenced by the improved discriminability in the ablation study.

At inference time, BGC forecasting relies exclusively on CGM data, ensuring a realistic, deployable setting. 
By decoupling subject characterization from glucose prediction, \SCGP{} exploits multimodal information when available, while maintaining robustness and practicality in real-world scenarios.

We also investigated identification accuracy using combinations of these features and CGM data. Although adding CGM improves performance, it would hinder the architecture's ability to operate with CGM-only data at inference.

\begin{table}[t]
\centering
\caption{Feature ablation highlighting subject discriminability based on different inputs on the OhioT1DM and DiaTrend datasets\textsuperscript{*}}
\label{tab:abl_ext_features}
\resizebox{\linewidth}{!}{%
\begin{tabular}{lccccccc}
\toprule
\multirow{2}{*}{\textbf{Features}} & \multicolumn{2}{c}{\textbf{Top-1 Acc}} & \multicolumn{2}{c}{\textbf{Top-3 Acc}} & \multicolumn{2}{c}{\textbf{Top-5 Acc}} \\
\cmidrule(lr){2-3} \cmidrule(lr){4-5} \cmidrule(lr){6-7}
 & OhioT1DM & DiaTrend & OhioT1DM & DiaTrend& OhioT1DM & DiaTrend \\
\midrule
Bolus & 29\% & 41\% & 58\% & 66\% & 75\% & 79\% \\
Carbs & 24\% & 35\% & 53\% & 60\% & 71\% & 74\% \\
Basal & 43\% & 75\% & 71\% & 95\% & 81\% & 98\% \\
\midrule
Bolus + Carbs & 49\% & 54\% & 74\% & 76\% & 86\% & 85\% \\
Bolus + Basal & 55\% & 82\% & 79\% & \textbf{98\%} & 89\% & \textbf{99\%} \\
Carbs + Basal & 49\% & 83\% & 76\% & 97\% & 85\% & \textbf{99\%} \\
\midrule
Bolus + Carbs + Basal & \textbf{65\%} & \textbf{84\%} & \textbf{86\%} & \textbf{98\%} & \textbf{93\%} & \textbf{99\%} \\
\bottomrule
\end{tabular}
}
\\
 \vspace{0.3em}
 \raggedright{
 \centering
 \scriptsize{\textsuperscript{*} In \textbf{bold} the best score.}\\
 }
\end{table}

\section{Conclusions}
In this work, we proposed \SCGP{}, a framework for personalized blood glucose forecasting that explicitly separates subject characterization from glucose dynamics modeling. By learning a compact subject-specific representation from contextual and treatment-related signals and using it to condition glucose prediction, our approach provides a structured and interpretable form of personalization beyond implicit adaptation strategies.

The \SCGP{} architecture is designed to align with realistic clinical deployment scenarios. During an initial observation phase, individual behavioral and physiological patterns are summarized into a persistent subject signature. Once learned, this representation can be retained and exploited to personalize BGC forecasts based solely on CGM data at inference time, making the approach suitable for real-world scenarios where auxiliary signals may be unavailable or unreliable.

Experiments on two large-scale benchmark datasets show that this design improves key aspects (i.e., time gain and adverse events detection), maintaining clinical reliability. These results highlight explicit subject conditioning as an effective personalization strategy under realistic clinical constraints.

\section*{Acknowledgment}
This work was funded by the National Plan for NRRP Complementary Investments (PNC, established with the decree-law 6 May 2021, n. 59, converted by law n. 101 of 2021) in the call for the funding of research initiatives for technologies and innovative trajectories in the health and care sectors (Directorial Decree n. 931 of 06-06-2022) - project n. PNC0000003 - AdvaNced Technologies for Human-centrEd Medicine (project acronym: ANTHEM)~\footnote{\url{https://fondazioneanthem.it/}}. This work reflects only the authors’ views and opinions, neither the Ministry for University and Research nor the European Commission can be considered responsible for them.

\bibliographystyle{IEEEtran}
\bibliography{bibliography}
\end{document}